\documentclass[letterpaper]{article} 
\usepackage{aaai24}  
\usepackage{times}  
\usepackage{helvet}  
\usepackage{courier}  
\usepackage[hyphens]{url}  
\usepackage{graphicx} 
\usepackage{natbib}  
\usepackage{caption} 
\usepackage{algorithm}
\usepackage{algorithmic}

\usepackage{booktabs}
\usepackage{enumitem}   
\usepackage{colortbl}  %
\usepackage{xcolor}
\usepackage{multirow}
\usepackage{amssymb}

\usepackage{newfloat}
\usepackage{listings}
\DeclareCaptionStyle{ruled}{labelfont=normalfont,labelsep=colon,strut=off} 
\floatstyle{ruled}
\newfloat{listing}{tb}{lst}{}
\floatname{listing}{Listing}
\title{VadCLIP: Adapting Vision-Language Models for Weakly Supervised \\Video Anomaly Detection}
\author{
    Peng Wu\textsuperscript{\rm 1}, Xuerong Zhou\textsuperscript{\rm 1}, Guansong Pang\textsuperscript{\rm 2}\footnote{Corresponding Authors}, Lingru Zhou\textsuperscript{\rm 1}, Qingsen Yan\textsuperscript{\rm 1},\\ Peng Wang\textsuperscript{\rm 1}$^\ast$, Yanning Zhang\textsuperscript{\rm 1}}
\affiliations{
    \textsuperscript{\rm 1}ASGO, School of Computer Science, Northwestern Polytechnical University, China\\
    \textsuperscript{\rm 2}School of Computing and Information Systems, Singapore Management University, Singapore\\

    \{xdwupeng, zxr2333\}@gmail.com, gspang@smu.edu.sg, \{lingruzhou, yqs\}@mail.nwpu.edu.cn,\\
    \{peng.wang, ynzhang\}@nwpu.edu.cn
}

\usepackage{bibentry}

\begin{document}

\maketitle

\begin{abstract}
The recent contrastive language-image pre-training (CLIP) model has shown great success in a wide range of image-level tasks, revealing remarkable ability for learning powerful visual representations with rich semantics. An open and worthwhile problem is efficiently adapting such a strong model to the video domain and designing a robust video anomaly detector. In this work, we propose VadCLIP, a new paradigm for weakly supervised video anomaly detection (WSVAD) by leveraging the frozen CLIP model directly without any pre-training and fine-tuning process. Unlike current works that directly feed extracted features into the weakly supervised classifier for frame-level binary classification, VadCLIP makes full use of fine-grained associations between vision and language on the strength of CLIP and involves dual branch. One branch simply utilizes visual features for coarse-grained binary classification, while the other fully leverages the fine-grained language-image alignment. With the benefit of dual branch, VadCLIP achieves both coarse-grained and fine-grained video anomaly detection by transferring pre-trained knowledge from CLIP to WSVAD task. We conduct extensive experiments on two commonly-used benchmarks, demonstrating that VadCLIP achieves the best performance on both coarse-grained and fine-grained WSVAD, surpassing the state-of-the-art methods by a large margin. Specifically, VadCLIP achieves 84.51\% AP and 88.02\% AUC on XD-Violence and UCF-Crime, respectively. Code and features are released at \textcolor{blue}{https://github.com/nwpu-zxr/VadCLIP}.
\end{abstract}

\section{Introduction}
In recent years, weakly supervised video anomaly detection (WSVAD, VAD) has received growing concerns due to its broad application prospects. For instance, with the aid of WSVAD, it is convenient to develop more powerful intelligent video surveillance systems and video content review systems. In WSVAD, the anomaly detector is expected to generate frame-level anomaly confidences with only video-level annotations provided. The majority of current research in this field follows a systematic process, wherein the initial step is to extract frame-level features using pre-trained visual models, e.g., C3D~\cite{tran2015learning, sultani2018real}, I3D~\cite{carreira2017quo, wu2020not}, and ViT~\cite{dosovitskiy2020image, li2022self}, followed by feeding these features into multiple instance learning (MIL) based binary classifiers for the purpose of model training, and the final step is to detect abnormal events based on predicted anomaly confidences. Despite their simple schemes and promising results, such a classification-based paradigm fails to take full advantage of cross-modal relationships, e.g, vision-language associations.

\begin{figure}[t]
  \centering
  \includegraphics[width=0.8\linewidth]{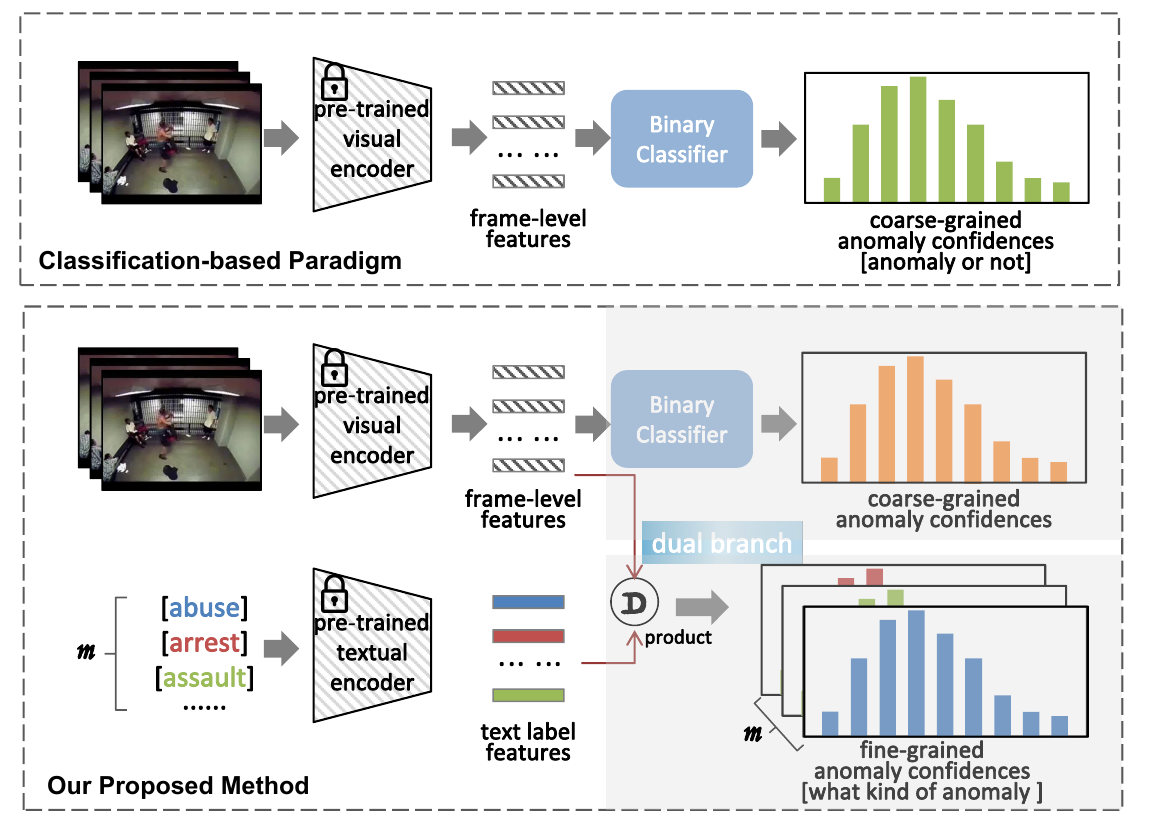}
  \caption{Comparisons of different paradigms for WSVAD.}
 
  \label{comparison}
  \vspace{-0.3cm} 
\end{figure}

During the past two years, we have witnessed great progress in the development of vision-language pre-training (VLP) models~\cite{kim2021vilt, jia2021scaling, wang2021simvlm, chen2023vlp}, e.g., CLIP~\cite{radford2021learning}, for learning more generalized visual representations with semantic concepts. The main idea of CLIP is to align images and texts by contrastive learning, that is, pull together images and matched textual descriptions while pushing away unmatched pairs in the joint embedding space. Thanks to hundreds of million noisy image-text pairs crawled from the web, such models pre-trained at a large scale really demonstrate their strong representation learning as well as associations between vision and language. In view of the breakthrough performance of CLIP, recently, building task-specific models on top of CLIP is becoming emerging research topics and applied to a broad range of vision tasks, and these models achieve unprecedented performance. 

Although CLIP and its affiliated models demonstrate the great potential on various vision tasks, these methods mainly focus on the image domain. Therefore, how to efficiently adapt such a model learned from image-text pairs to more complex video anomaly detection task under weak supervision deserves a thorough exploration. Recently, a few works~\cite{joo2023clip, lv2023unbiased} attempt to make use of the learned knowledge of CLIP, however, these methods limit their scope to directly using visual features extracted from the image encoder of CLIP,  and neglect to exploit semantic relationships between vision and language.

In order to make effective use of generalized knowledge and enable CLIP to reach its full potential on WSVAD task, based on the characteristics of WSVAD, there are several critical challenges that need to be addressed. First, it is vital to explore ways to capture contextual dependencies across time. Second, it is essential to determine how to harness learned knowledge and the visual-language connections. Third, it is crucial to maintain optimal CLIP performance under weak supervision.

In this work, we propose a novel paradigm based on CLIP for WSVAD, which is dubbed as \textbf{VadCLIP}. VadCLIP consists of several components to overcome the above challenges. Specifically, \textbf{for the first challenge}, we present a local-global temporal adapter (LGT-Adapter), which is a lightweight module for video temporal relation modeling. LGT-Adapter involves two components, i.e., local temporal adapter and global temporal adapter, wherein the former mainly captures local temporal dependencies with high efficiency, since in most cases the current events are highly related to the adjacent events, and the latter smooths feature information in a more holistic view with less parameters. \textbf{For the second challenge}, unlike current methods~\cite{joo2023clip, lv2023unbiased} that solely use visual features, we encourage VadCLIP to also leverage textual features to preserve learned knowledge as much as possible. As shown in Figure~\ref{comparison}, VadCLIP is devised as a dual-branch fashion, where one simply and directly utilizes visual features for binary classification (C-branch), while the other employs both visual and textual features for language-image alignment (A-branch). Moreover, such dual branch seamlessly achieves coarse-grained and fine-grained WSVAD~\cite{wu2022weakly}. For A-branch, we build bridge between videos and video-level textual labels. Moreover, we propose two prompt mechanisms~\cite{wu2023towards}, i.e., learnable prompt and visual prompt, to specify that the succinct text is about the video. Learnable prompt does not require extensive expert knowledge compared to the handcrafted prompt, effectively transfers pre-trained knowledge into the downstream WSVAD task. Visual prompt is inspired by that visual contexts can make the text more accurate and discriminate. Imagine that if there is a car in the video, two types of abnormal events of "car accident" and "fighting" would be more easily distinguished. Hence, In the visual prompt, we focus on anomaly information in videos and integrate these anomaly-focus visual contents from C-branch with textual labels from A-branch for automatic prompt engineering. Such a practice seamlessly creates connections between dual branch. \textbf{For the third challenge}, multiple instance learning (MIL)~\cite{sultani2018real, wu2020not} is the most commonly used method. For the language-visual alignments in A-branch, we introduce a MIL-Align mechanism, the core idea is to select the most matched video frames for each label to represent the whole video.

Note that during training, the weights of CLIP image and text encoders are kept fixed, and the gradients are back-propagated to optimise these learnable parameters of the devised adapter and prompt modules.

Overall, the contributions of our work are threefold:

\noindent\textbf{(1)} 
We present a novel diagram, i.e., VadCLIP, which involves dual branch to detect video anomaly in visual classification and language-visual alignment manners, respectively. With the benefit of dual branch, VadCLIP achieves both coarse-grained and fine-grained WSVAD. To our knowledge, VadCLIP is the first work to efficiently transfer pre-trained language-visual knowledge to WSVAD.

\noindent\textbf{(2)} 
We propose three non-vital components to address new challenges led by the new diagram. LGT-Adapter is used to capture temporal dependencies from different perspectives; Two prompt mechanisms are devised to effectively adapt the frozen pre-trained model to WSVAD task; MIL-Align realizes the optimization of alignment paradigm under weak supervision, so as to preserve the pre-trained knowledge as much as possible.

\noindent\textbf{(3)} 
We show that strength and effectiveness of VadCLIP on two large-scale popular benchmarks, and VadCLIP achieves state-of-the-art performance, e.g., it obtains unprecedented results of 84.51\% AP and 88.02\% AUC on XD-Violence and UCF-Crime respectively, surpassing current classification based methods by a large margin.

\section{Related Work}
\subsection{Weakly Supervised Video Anomaly Detection}
Recently, some researchers~\cite{zaheer2020claws, feng2021mist, wu2021weakly, chen2023mgfn} have proposed weakly supervised methods for VAD. Sultani et al.~\cite{sultani2018real} firstly proposed a deep multiple instance learning model, which considers a video as a bag and its multiple segments as instances. Then several follow-up works made effort to model temporal relations based on self-attention models and transformers. For example, Zhong et al.~\cite{zhong2019graph} proposed a graph convolutional network (GCN) based method to model the feature similarity and temporal consistency between video segments. Tian et al.~\cite{tian2021weakly} used a self-attention network to capture the global temporal context relationship of videos. Li et al.~\cite{li2022self} proposed a transformer based multi-sequence learning framework, and Huang et al.~\cite{huang2022weakly} proposed a transformer based temporal representation aggregation framework. Zhou et al.~\cite{zhou2023dual} presented a global and local multi-head self attention module for the transformer layer to obtain more expressive embeddings for capturing temporal dependencies in videos. The above methods only detect whether video frames are anomalous, on the contrary, Wu et al.~\cite{wu2022weakly} proposed a fine-grained WSVAD method, which distinguishes between different types of anomalous frames. More recently, the CLIP model has also attracted great attentions in the VAD community. Based on visual features of CLIP, Lv et al.~\cite{lv2023unbiased} proposed a new MIL framework called Unbiased MIL (UMIL) to learn unbiased anomaly features that improve WSVAD performance. Joo et al.~\cite{joo2023clip} proposed to employ visual features from CLIP to efficiently extract discriminative representations, and then model long- and short-range temporal dependencies and nominate the snippets of interest by leveraging temporal self-attention. All the above methods are based on the classification paradigm, which detect anomalous events by predicting the probability of anomalous frames. However, this classification paradigm does not fully utilize the semantic information of textual labels.

\subsection{Vision-Language Pre-training} 
Vision-language pre-training has achieved impressive progress over the past few years, which aims to learn the semantic correspondence between vision and language through pre-training on large-scale data. As one of the most representative works, CLIP has shown impressive performance on a range of vision-language downstream tasks, including image classification~\cite{zhou2022learning}, image captioning~\cite{mokady2021clipcap}, object detection~\cite{zhou2022detecting}, scene text detection~\cite{yu2023turning}, dense prediction~\cite{zhou2022zegclip,rao2022denseclip}, and so on. Recently, some follow-up works attempted to leverage the pre-trained models for video domains. For example, CLIP4Clip~\cite{luo2022clip4clip} transferred the knowledge of CLIP model to the video-text retrieval, some works~\cite{wang2021actionclip,lin2022frozen, ni2022expanding} attempted to take advantages of CLIP for video recognition, furthermore, CLIP is used to tackle the more complex video action localization task~\cite{nag2022zero, ju2022prompting}. More generally, Ju et al.~\cite{ju2022prompting} presented a simple yet strong baseline to efficiently adapt the pre-trained image-based visual-language model, and exploited its powerful ability for general video understanding. In this work, we deeply explore how to adapt pre-trained vision-language knowledge of CLIP from image-level into video-level downstream WSVAD efficiently.

\begin{figure*}[t]
  \centering
  \includegraphics[width=0.75\linewidth]{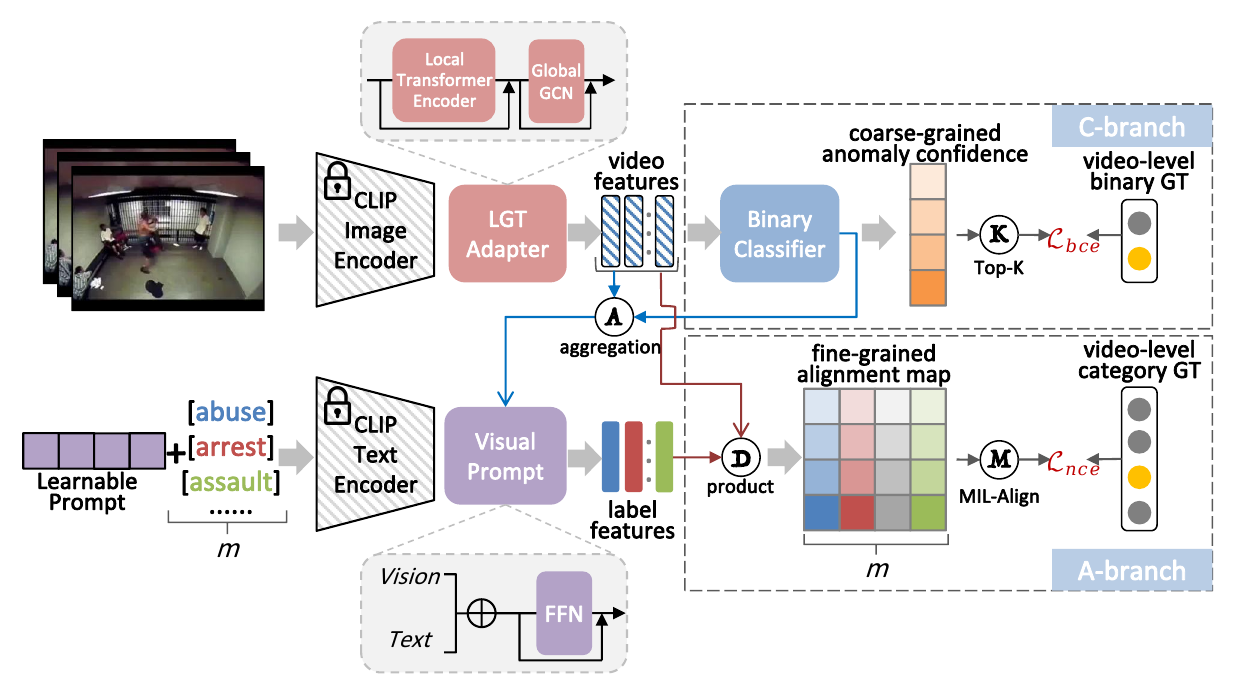}
  \caption{The framework of our proposed VadCLIP.}
  \label{framework}
\end{figure*}

\section{Method}
\subsection{Problem Definition}\label{sec:definition}
The WSVAD task supposes that only video-level labels are available during the training stage. Given a video $v$, if all frames of this video do not contain abnormal events, this video is defined as normal with the label $y=0$; Otherwise, if there is at least one frame contains abnormal events, this video is labeled as abnormal with the label $y=1$. The goal of WSVAD task is to train a detection model that is able to predict frame-level anomaly confidences while only video-level annotations are provided.

Previous works generally make use of pre-trained 3D convolutional models, e.g., C3D~\cite{tran2015learning} and I3D~\cite{carreira2017quo}, to extract video features, and then feed these features into MIL-based binary classifiers, such paradigms are referred as the classification-based paradigm in this paper. Recently, CLIP, as a large-scale language-vision pre-trained model, has revolutionized many fields in computer vision, and has shown great generalization capabilities across a wide range of downstream tasks. Inspired by CLIP, our work not only uses the image encoder of CLIP as the backbone to extract video features, but also attempts to utilize the text encoder of CLIP to take full advantage of the powerful associations between visual contents and textual concepts. Our work is demonstrated in Figure~\ref{framework}.

\subsection{Local and Global Temporal Adapter}\label{sec:lgt-adapter}
As we know, CLIP is pre-trained on large-scale image-text pairs crawled from the web. In this section, we investigate how to model temporal dependencies and bridge the gap between the image domain and video domain for CLIP. Meanwhile, it is also significant to learn long-range and short-range temporal dependencies for WSVAD task~\cite{zhou2023dual, wu2021learning}. From the perspective of the efficiency and receptive field, we design a new temporal modeling method compatible with local and global receptive field. 

\subsubsection{Local Module.}
To capture local temporal dependencies, we introduce a transformer encoder layer on top of frame-level features $X_{clip}\in\mathbb{R}^{n\times d}$ from the frozen image encoder of CLIP, where $n$ is the length of video, $d$ is the dimension size, which is set as 512 in this work. Note that this layer differs from the ordinary transformer encoder layer since it limits self-attention computation to local windows~\cite{liu2021swin} instead of the global scope. Specifically, we split frame-level features into equal-length and overlapping windows over temporal dimension, self-attention calculation is limited within each window, and no information exchange among windows. Such an operation possesses local receptive field like the convolution, and leads to the lower computation complexity.

\subsubsection{Global Module.}
To further capture global temporal dependencies, we introduce a lightweight GCN module following local module, we adopt GCN to capture global temporal dependencies due to its widespread adoption and proven performance in VAD~\cite{zhong2019graph, wu2020not, wu2021learning}. Following the setup in~\cite{zhong2019graph, wu2020not}, we use GCN to model global temporal dependencies from the perspective of feature similarity and relative distance, it can be summarized as follows,
\begin{equation}
  X_g = gelu\left(\left[Softmax\left(H_{sim}\right);Softmax\left(H_{dis}\right)\right]X_{l}W\right)
\end{equation}
where $H_{sim}$ and $H_{dis}$ are the adjacency matrices, the Softmax normalization is used to ensure the sum of each row of  $H_{sim}$ and $H_{dis}$ equals to one. $X_{l}$ is the frame-level video feature obtained from local module, $W$ is the only one learnable weight that is used to transform the feature space, this setup demonstrates the lightweight of global module. 

\textbf{Feature similarity branch} is designed to generated a similarity relationship adjacency matrix for GCN. We use the frame-wise cosine similarity to calculate the adjacency matrix $H_{sim}$, which is presented as follows,
\begin{equation}
  H_{sim} = \frac{X_{l}X_{l}^{\top}}{\left\|X_{l}\right\|_{2}\cdot\left\|X_{l}\right\|_{2}}
  \label{con:4}
\end{equation}
we also use the thresholding operation to filter weak relations~\cite{wu2020not}. 

\textbf{Position distance branch} is used to capture long-range dependencies based on positional distance between each two frames. The proximity adjacency matrix is shown as follows:
\begin{equation}
H_{dis}(i,j)=\frac{-|i-j|}{\sigma }\label{con:position}
\end{equation}
the proximity relation between $i^{th}$ and $j^{th}$ frames only determined by their relative temporal position. $\sigma$ is a hyper-parameter to control the range of influence of distance relation. Both local transformer and GCN layer employ residual connection to prevent feature over-smoothing.

\subsection{Dual Branch and Prompt}\label{sec:dual branch}
\subsubsection{Dual Branch.}
Unlike other previous WSVAD works, our VadCLIP contains dual branch, more precisely, in addition to the traditional binary classification branch (C-Branch), we also introduce a novel video-text alignment branch, dubbed as A-Branch. Specifically, after temporal modeling, the video feature $X_g$ is fed into a fully connected (FC) layer to obtain the final video feature $X \in \mathbb{R}^{n \times d}$. In C-Branch, we feed $X$ into a binary classifier that contains a feed-forward network (FFN) layer, an FC layer and a Sigmoid activation to obtain the anomaly confidence $A\in\mathbb{R}^{n \times 1}$.
\begin{equation}
A= Sigmoid\left(FC\left(FFN\left(X\right)+X\right)\right)
\label{con:binary class}
\end{equation}
In A-Branch, textual labels, e.g., abuse, riot, fighting, etc, are no longer encoded as one-hot vectors, on the contrary, they are encoded into class embeddings using the text encoder of CLIP, we leverage the frozen pre-trained text encoder of CLIP throughout, as the text encoder can provide language knowledge prior for video anomaly detection. Then we calculate the match similarities between class embeddings and frame-level visual features to obtain the alignment map $M\in\mathbb{R}^{n\times m}$, where $m$ is the number of text labels, such a setup is similar to that of CLIP. In A-Branch, each input text label represents a class of abnormal events, thus naturally achieving fine-grained WSVAD.

\subsubsection{Learnable Prompt.}
In WSVAD, text labels are words or phrases, which are too succinct to summarize abnormal events very well. To learn robust transferability of text embedding, we take inspirations from CoOp~\cite{zhou2022learning}, and add the learnable prompt to original class embeddings. Concretely, the original text labels are first transformed into class tokens through CLIP tokenizer, i.e., $t_{init} = Tokenizer(Label)$, 
where $Label$ is the discrete text label, e.g., fighting, shooting, road accident, etc. Then we concatenate $t_{init}$ with the learnable prompt $\{c_1, ... , c_l\}$ that contains $l$ context tokens to form a complete sentence token, thus the input of text encoder is presented as follows:
\begin{equation}
  t_p = \{c_1, ..., t_{init}, ..., c_l\}
  \label{con:token_in}
\end{equation}
here we place the class token at the middle of a sequence. Then this sequence token is added to the positional embedding to obtain positional information, and finally, the text encoder of CLIP takes as input $t_p$ and generates class embedding $t_{out}\in\mathbb{R}^{d}$.
\subsubsection{Anomaly-Focus Visual Prompt.}
In order to further improve the representation ability of text labels for abnormal events, we investigate how to use visual contexts to refine the class embedding, since visual contexts can make the succinct text labels more accurate. To this end, we propose an anomaly-focus visual prompt, which focuses on the visual embeddings in abnormal segments, and aggregate these embeddings as the video-level prompt for class embeddings. We first use the anomaly confidence $A$ obtained from C-Branch as the anomaly attention, then compute the video-level prompt by the dot product of anomaly attention and video feature $X$, which is presented as follows,
\begin{equation}
  V = Norm\left(A^{\top}X\right)
  \label{con:attention}
\end{equation}
where $Norm$ is the normalization, and $V\in\mathbb{R}^{d}$ is the anomaly-focus visual prompt. We then add $V$ to the class embedding $t_{out}$ and obtain the final instance-specific class embedding $T$ by a simple FFN layer and a skip connection. 
\begin{equation}
  T = FFN\left(ADD\left(V,t_{out}\right)\right)+t_{out}
  \label{con:visual_prompt}
\end{equation}
where $ADD$ is the element-wise addition. Such a implementation allows class embeddings to extract the related visual context from videos. 

With $X$ and $T$ in hands, we calculate the match similarities between all class embeddings and frame-level visual features to obtain
the alignment map $M$.

\subsection{Objective Function}\label{sec:objective}
For C-Branch, we follow previous works~\cite{wu2020not} and use Top-K mechanism to select $K$ high anomaly confidences in both abnormal and normal videos as the video-level predictions. Then we use the binary cross entropy between video-level predictions and ground-truth to compute classification loss $\mathcal{L}_{bce}$.

For A-Branch, we are confronted with new challenges: 1) there is no anomaly confidence; 2) facing multi-classes instead of binary classes. To address this dilemma, we propose the MIL-Align mechanism which is similar to vanilla MIL. Specifically, we consider the align map $M$ since it expresses the similarity between frame-level video features and all class embeddings. For each row, we select top $K$ similarities and compute the average to measure the alignment degree between this video and the current class. Then we can obtain a vector $S=\{s_1,...,s_m\}$ that represents the similarity between this video and all classes. We hope the video and its paired textual label emit the highest similarity score among others. To achieve this, the multi-class prediction is firstly computed as follows,
\begin{equation}
  p_i = \frac{exp\left(s_i/\tau\right)}{\sum_j{exp\left(s_j/\tau\right)}}
  \label{con:nce}
\end{equation}
where $p_i$ is the prediction with respect to the $i^{th}$ class, and $\tau$ refers to the temperature hyper-parameter for scaling. Finally, the alignment loss $\mathcal{L}_{nce}$ can be computed by the cross entropy.

In addition to classification loss $\mathcal{L}_{bce}$ and alignment loss $\mathcal{L}_{nce}$, we also introduce a contrastive loss to slightly push the normal class embedding and other abnormal class embeddings away, here we first calculate cosine similarity between normal class embedding and other abnormal class embeddings, and then compute the contrastive loss $\mathcal{L}_{cts}$ as follows,
\begin{equation}
 \mathcal{L}_{cts} = \sum_j{max\left(0,\frac{t_{n}^{\top}t_{aj}}{\left\|t_{n}\right\|_{2}\cdot\left\|t_{aj}\right\|_{2}}\right)}
\label{con:constractive}
\end{equation}
where $t_n$ is the normal class embedding, and $t_a$ is abnormal class embeddings.

Overall, the final total objective of VadCLIP is given by:
\begin{equation}
  \mathcal{L} = \mathcal{L}_{bce}+\mathcal{L}_{nce}+\lambda\mathcal{L}_{cts}
  \label{eq:loss}
\end{equation}

\subsection{Inference}\label{sec:inference}
VadCLIP contains dual branch that enables itself to address both fine-grained and coarse-grained WSVAD tasks. In regard to fine-grained WSVAD, we follow previous works~\cite{wu2022weakly} and utilize a thresholding strategy on alignment map $M$ to predict anomalous events. In regard to coarse-grained WSVAD, there are two ways to compute the frame-level anomaly degree. The first one is to directly use the anomaly confidences in C-Branch, the second one is to use the alignment map in A-Branch, specifically, subtracting the similarities between videos and the normal class by one is the anomalous degree. Finally, we select the best of these two ways for computing the frame-level anomaly degree.

\section{Experiments}
\subsection{Datasets and Evaluation Metrics}
\subsubsection{Datasets.} 
We conduct experiments on two popular WSVAD datasets, i.e., UCF-Crime and XD-Violence. Notably, training videos only have video-level labels on both datasets. 

\subsubsection{Evaluation Metrics.}
For coarse-grained WSVAD, we follow previous works, and utilize the frame-level Average Precision (AP) for XD-Violence, and frame-level AUC and the AUC of anomaly videos (termed as AnoAUC) for UCF-Crime. 
For fine-grained WSVAD, we follow the standard evaluation protocol in video action detection and use the mean Average Precision (mAP) values under different intersection over union (IoU) thresholds. In this work, we use IoU thresholds ranging from 0.1 to 0.5 with a stride of 0.1 to compute mAP values. Meanwhile, we also report an average of mAP (AVG). Note that we only compute mAP on the abnormal videos in the test set. 

\subsection{Implementation Details}
For network structure, frozen image and text encoders are adopted from pre-trained CLIP (ViT-B/16). FFN is a standard layer from Transformer, and ReLU is replaced with GELU. 
For hyper-parameters, 
we set $\sigma$ in Eq.\ref{con:position} as 1, $\tau$ in Eq.\ref{con:nce} as 0.07, and the context length $l$ as 20. For window length in LGT-Adapter, we set it as 64 and 8 on XD-Violence and UCF-Crime, respectively. For $\lambda$ in Eq.\ref{eq:loss}, we set it as $1\times 10^{-4}$ and $1\times 10^{-1}$ on XD-Violence and UCF-Crime, respectively.
For model training, VadCLIP is trained on a single NVIDIA RTX 3090 GPU using PyTorch. We use AdamW 
as the optimizer with batch size of 64. On XD-Violence, the learning rate and total epoch are set as $2 \times 10^{-5}$ and 20, respectively, and on UCF-Crime, the learning rate and total epoch are set as $1 \times 10^{-5}$ and 10, respectively. 

\subsection{Comparison with State-of-the-Art Methods}
VadCLIP can simultaneously realize coarse-grained and fine-grained WSVAD, therefore we present the performance of VadCLIP and compare it with several state-of-the-art methods on coarse-grained and fine-grained WSVAD tasks. For the sake of fairness, \textit{all comparison methods use the same visual features extracted from CLIP as VadCLIP.}

\subsubsection{Coarse-grained WSVAD Results.} 
We show comparison results in Tables \ref{tabxd} and \ref{tabucf}. Here Ju et al.~\cite{ju2022prompting} is a CLIP-based work for action recognition, which is significantly inferior to our method. Such results demonstrate challenges on WSVAD task, and also show the strength of our method with respect to Ju et al.~\cite{ju2022prompting} for the specific WSVAD task. Besides,
we found that VadCLIP significantly outperforms both semi-supervised methods and classification-based weakly supervised methods on two commonly-used benchmarks and across all evaluation metrics. More precisely, VadCLIP attains 84.51\% AP and 82.08\% AUC on XD-Violence and UCF-Crime, respectively, a new state-of-the-art on both datasets. By comparison, VadCLIP achieves an absolute gain of 2.3\% and 2.1\% in terms of AP over the best competitors CLIP-TSA~\cite{joo2023clip} and DMU~\cite{zhou2023dual} on XD-Violence, and on UCF-Crime, VadCLIP also outperforms them by 0.4\% and 1.3\% in terms of AUC. More importantly, among all comparison methods, AVVD~\cite{wu2022weakly} uses fine-grained class labels exactly, and it only achieves 78.10\% AP and 82.45\% AUC on XD-Violence and UCF-Crime, respectively, which lags behind VadCLIP by a large margin. Such a result shows simply using fine-grained labels cannot lead to performance gains, since excessive inputs of label increases the difficulty of binary classification. The performance advantage of VadCLIP is partially attributable to the vision-language associations, since all comparison baselines use the same visual features as VadCLIP.

\subsubsection{Fine-grained WSVAD Results.} 
For fine-grained WSVAD task, we compare VadCLIP with previous works AVVD and  Sultani et al.~\cite{sultani2018real, wu2022weakly} in Tables~\ref{tabfine-xd} and~\ref{tabfine-ucf}. Here AVVD is the first work to propose the fine-grained WSVAD, and we re-implement it with visual features of CLIP, then we also fine-tune Sultani et al. based on the setup in AVVD for adapting fine-grained WSVAD. As we can see, the fine-grained WSVAD is a more challenging task with respect to coarse-fined  WSVAD since the former needs to consider both multi-category classification accuracy and detection segment continuity. On this task, VadCLIP is also clearly superior to these excellent comparison methods on both XD-Violence and UCF-Crime datasets. For instance, On XD-Violence, VadCLIP achieves a performance improvement of 13.1\% and 4.5\% in terms of AVG compared to Sultani et al. and AVVD.

\subsection{Ablation Studies}
Extensive ablations are carried out on XD-Violence dataset. Here we choose the similarity map to compute the frame-level anomaly degree for coarse-grained WSVAD.

\begin{table}[t]
  \centering
  
  \resizebox{\linewidth}{!}{
  \begin{tabular}{c|lc}
    \toprule
    Category&Method&AP(\%)\\
    \midrule
     & SVM baseline &50.80 \\
     Semi & OCSVM~\cite{scholkopf1999support} &28.63\\ 
     & Hasan et al.~\cite{hasan2016learning} & 31.25\\ 
    \midrule
    &Ju et al.~\cite{ju2022prompting}& 76.57\\
    &Sultani et al.~\cite{sultani2018real}& 75.18\\
    &Wu et al.~\cite{wu2020not} &80.00\\ 
    &RTFM~\cite{tian2021weakly} &78.27\\ 
    Weak&AVVD~\cite{wu2022weakly}&78.10\\ 
    &DMU~\cite{zhou2023dual} & 82.41\\
    &CLIP-TSA~\cite{joo2023clip} & 82.17\\
    &\cellcolor{gray!20}\textbf{VadCLIP (Ours)} & \cellcolor{gray!20}\textbf{84.51} \\
  \bottomrule
\end{tabular}}

\caption{Coarse-grained comparisons on XD-Violence.}
\label{tabxd}
\end{table}

\begin{table}[!t]
  \centering
  
  \label{tabucf}
  \resizebox{0.9\linewidth}{!}{
  \begin{tabular}{c|lcc}
    \toprule
    Category&Method&AUC(\%)&Ano-AUC(\%)\\
    \midrule
     & SVM baseline &50.10 & 50.00\\
     Semi & OCSVM~\shortcite{scholkopf1999support} &63.20 & 51.06\\
     & Hasan et al.~\shortcite{hasan2016learning} & 51.20& 39.43\\
    \midrule
    &Ju et al.~\shortcite{ju2022prompting}& 84.72 & 62.60\\
    &Sultani et al.~\shortcite{sultani2018real}& 84.14& 63.29\\
    &Wu et al.~\shortcite{wu2020not} &84.57 & 62.21\\
    &AVVD~\shortcite{wu2022weakly} & 82.45&60.27 \\
    &RTFM~\shortcite{tian2021weakly} &85.66 & 63.86\\
    Weak&DMU~\shortcite{zhou2023dual} & 86.75& 68.62\\
    &UMIL~\shortcite{lv2023unbiased} & 86.75& 68.68\\
    &CLIP-TSA~\shortcite{joo2023clip} & 87.58& N/A \\
    &\cellcolor{gray!20}\textbf{VadCLIP (Ours)} & \cellcolor{gray!20}\textbf{88.02} &\cellcolor{gray!20}\textbf{70.23} \\
  \bottomrule
\end{tabular}}

\caption{Coarse-grained comparisons on UCF-Crime.}
\label{tabucf}
\end{table}

\begin{table}[!t]
  \centering
  \resizebox{\linewidth}{!}{
  \begin{tabular}{l|cccccc}
    \toprule
    \multirow{2}{*}{Method}  & \multicolumn{6}{c}{mAP@IOU(\%)}\\
    ~ & 0.1 & 0.2 & 0.3 & 0.4 & 0.5 & AVG  \\
    \hline
    Random Baseline  & 1.82 & 0.92 & 0.48 & 0.23 & 0.09 & 0.71  \\
    Sultani et al.~\shortcite{sultani2018real}& 22.72 & 15.57 & 9.98 & 6.20 & 3.78& 11.65 \\
    AVVD~\shortcite{wu2022weakly} & 30.51 & 25.75 & 20.18 &14.83 & 9.79& 20.21  \\
    \rowcolor{gray!20}\textbf{VadCLIP (Ours)} & \textbf{37.03} & \textbf{30.84} & \textbf{23.38} & \textbf{17.90} & \textbf{14.31} & \textbf{24.70} \\
    \bottomrule
  \end{tabular}}
  \caption{Fine-grained comparisons on XD-Violence.}
  \label{tabfine-xd}
\end{table}

\begin{table}[!t]
  \centering
  \resizebox{\linewidth}{!}{
  \begin{tabular}{l|cccccc}
    \toprule
    \multirow{2}{*}{Method}  & \multicolumn{6}{c}{mAP@IOU(\%)}\\
    ~ & 0.1 & 0.2 & 0.3 & 0.4 & 0.5 & AVG  \\
    \hline
    Random Baseline  & 0.21 & 0.14 & 0.04 & 0.02 & 0.01 & 0.08  \\
    Sultani et al.~\shortcite{sultani2018real}& 5.73 & 4.41 & 2.69 & 1.93 & 1.44 & 3.24 \\
    AVVD~\shortcite{wu2022weakly} & 10.27 & 7.01 & 6.25 & 3.42 & \textbf{3.29}& 6.05  \\
    \rowcolor{gray!20}\textbf{VadCLIP (Ours)} & \textbf{11.72} & \textbf{7.83} & \textbf{6.40} & \textbf{4.53} & 2.93 & \textbf{6.68} \\
    \bottomrule
  \end{tabular}}
  \caption{Fine-grained comparisons on UCF-Crime.}
  \label{tabfine-ucf}
\end{table}

\subsubsection{Effectiveness of LGT-Adapter.} \label{sec:LGT}
As shown in Table~\ref{tabLGT-Adapter}, firstly, without the assistance of LGT-Adapter for temporal modeling, the baseline model only achieves 72.22\% AP and 15.64\% AVG, this results in a considerably drop of 12.3\% AP and 9.1\% AVG. Secondly, only using global transformer encoder layer, local transformer encoder layer or GCN layer gets clear performance boosts, especially in terms of AP, which convincingly indicates transformer encoder and GCN both can efficiently capture temporal dependencies by means of the self-attention mechanism across video frames. Thirdly, the combination of global transformer encoder and GCN yields the slightly improved performance in terms of AP (+0.4\%) over the combination of local transformer encoder and GCN, while the latter achieves significantly better performance in terms of AVG (+3.9\%). 
We also attempt a combination of local Transformer encoder and global Transformer encoder, which results in significant performance degradation in terms of AP listed in the $5^{th}$ row. The possible reason is that, compared to Transformer, GCN can be regarded as a lightweight variant, 
and fewer parameters prevent learned knowledge of CLIP from being affected during the transfer process.
Therefore, local transformer encoder and GCN are the optimum combination, which can capture different range temporal dependencies.

\begin{table}[t]
  \centering

  \resizebox{\linewidth}{!}{
    \begin{tabular}{l|cc}
    \toprule
    Method & \multicolumn{1}{l}{AP(\%)} & \multicolumn{1}{l}{AVG(\%)} \\
    \hline
    Baseline (w/o temporal modeling)& 72.22 & 15.64 \\
    Global TF-Encoder& 82.54 & 16.76 \\
    Local TF-Encoder& 81.18 & 18.41 \\
    Only GCN& 81.56 & 23.31 \\
    Local TF-Encoder+ Global TF-Encoder& 79.91 & 19.78\\
    Global TF-Encoder+GCN & \textbf{84.87} & 20.84 \\
    Local TF-Encoder+GCN (\textbf{LGT-Adapter}) & 84.51  & \textbf{24.70}  \\
    \bottomrule
    \end{tabular}}%
    \caption{Effectiveness of LGT-Adapter.}
    \label{tabLGT-Adapter}%
\end{table}%

\begin{table}[t]
  \centering

  \resizebox{0.95\linewidth}{!}{
    \begin{tabular}{cccc|c}
    \toprule
    C-Branch & A-Branch & L-Prompt & V-Prompt & \multicolumn{1}{c}{AP(\%)} \\
    \hline
    $\surd$ & & & & 80.53  \\
    \hline
     &$\surd$ & & & 68.15  \\
     $\surd$&$\surd$ & & & 75.03  \\
    $\surd$&$\surd$ & $\surd$& & 78.27  \\
    $\surd$&$\surd$ & & $\surd$& 82.35  \\
    $\surd$&$\surd$ &$\surd$ &$\surd$ & \textbf{84.51}  \\
    \bottomrule
    \end{tabular}}%
    \caption{Effectiveness of dual branch.}
    \label{tabdual branch}%
\end{table}%

\begin{table}[!t]
  \centering
  \resizebox{\linewidth}{!}{
    \begin{tabular}{c|ll}
    \toprule
     & \multicolumn{1}{l}{AP(\%)} & \multicolumn{1}{l}{AVG(\%)} \\
    \hline
    Hand-crafted Prompt  & 81.06 (\textcolor{green}{-3.46}) & 22.46 (\textcolor{green}{-2.24}) \\
    Learnable-Prompt & \textbf{84.51}  & \textbf{24.70} \\
    \hline
    Average-Frame Visual Prompt & 81.34 (\textcolor{green}{-3.17}) & 21.57 (\textcolor{green}{-3.13}) \\
    Anomaly-Focus Visual Prompt & \textbf{84.51}  & \textbf{24.70} \\
    \bottomrule
    \end{tabular}}%
    \caption{Effectiveness of prompt.}
     \label{tabprompt}%
     \vspace{-0.3cm}
\end{table}%

\begin{figure}[t]
  \centering
  \includegraphics[width=1.0\linewidth]{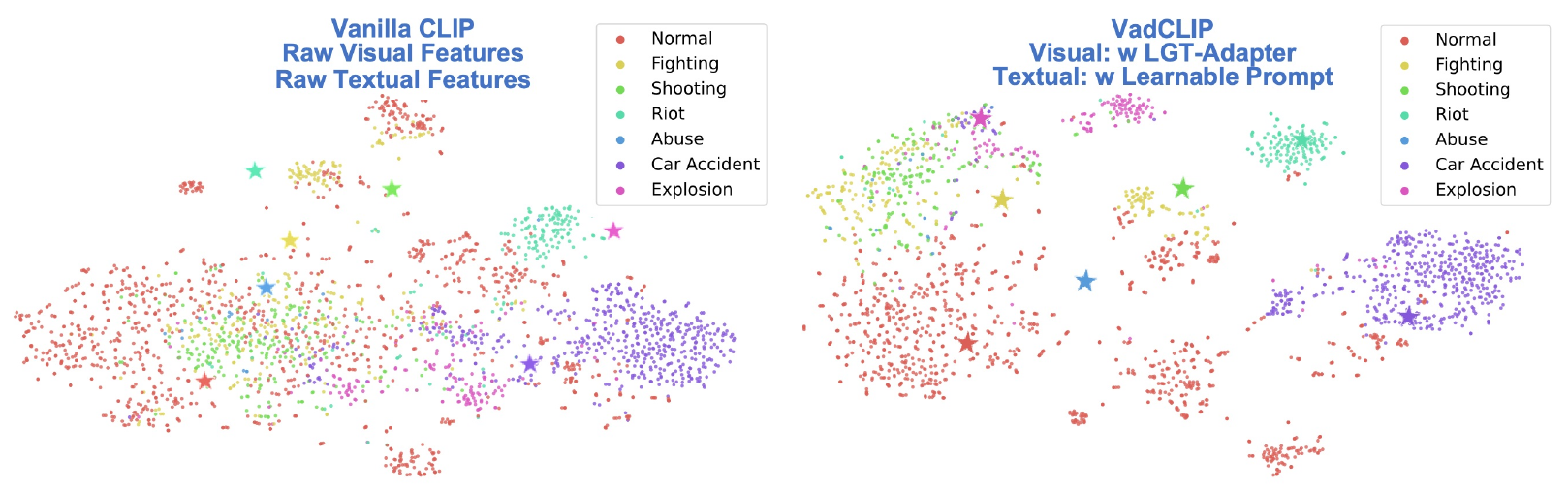}
  \caption{t-SNE visualizations for XD-Violence. Left: Raw CLIP features; Right: VadCLIP features.} 
  \label{tsne}
\end{figure}

\begin{figure}[t]
  \centering
  \includegraphics[width=1.0\linewidth]{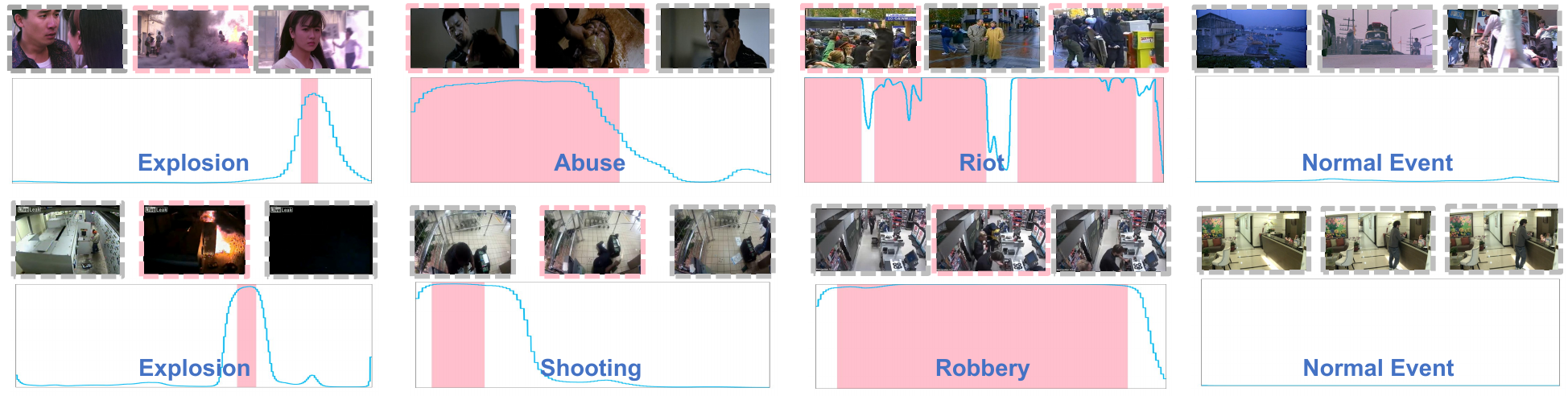}
  \caption{Qualitative results of coarse-grained WSVAD. }
  \label{coarse visualization}
\end{figure}

\subsubsection{Effectiveness of Dual Branch.} 
As shown in Table~\ref{tabdual branch}, our method with only C-Branch belongs to the classification-based paradigm, and can compete current state-of-the-art methods on XD-Violence. On the other hand, our method with only A-Branch achieves unsatisfactory performance in terms of AP since it is mainly focus on fine-grained WSVAD. With the assistance of coarse-grained classification on feature optimization in C-Branch, A-Branch obtains a leap of about 7\% AP improvement. By further adding the learnable prompt and visual prompt that are ad-hoc designs in A-Branch, we notice that a consistent performance improvement can be achieved, leading to a new state-of-the-art. These results clearly show dual branch that contains coarse-grained classification paradigm and fine-grained alignment paradigm can boost the 
performance by leveraging the complementary of different granularity.

\subsubsection{Effectiveness of Prompt.} 
As shown in Table~\ref{tabprompt}, using hand-crafted prompt results in a drop of 3.5\% AP and 2.2\% AVG, demonstrating that the learnable prompt has better potential for adapting pre-trained knowledge from the large language-vision model to WSVAD task. Furthermore, simply using the average of frame-level features in visual prompt~\cite{ni2022expanding} produces a drop of 3.2\% AP and 3.1\% AVG, such results show focusing on abnormal snippets in the video can support VadCLIP to obtain more accurate instance-specific text representations, which boosts the ability of video-language alignment that is useful for WSVAD task. We refer readers to \textbf{\textit{supplement materials}}\footnote{https://arxiv.org/abs/2308.11681} for more ablation studies and qualitative visualizations.

\subsection{Qualitative Analyses}
\subsubsection{Feature Discrimination Visualization.} 
We visualize the feature distribution by using t-SNE 
for XD-Violence, and present results in Figure~\ref{tsne}, where star icons denote textual label features. As we can see, although CLIP has learned generalized capacities based on image-text pairs, such capacities still cannot allow it to effectively distinguish different categories for WSVAD due to intrinsic problems on WSVAD task. After specialized optimization by VadCLIP, these visual features have more distinguishable boundaries and also surround the corresponding text class features.

\subsubsection{Coarse-grained Qualitative Visualization.} 
We illustrate the qualitative visualizations of coarse-grained WSVAD in Figure~\ref{coarse visualization}, where the blue curves represent the anomaly prediction, and the pink regions correspond to the ground-truth abnormal temporal location. As we can see, VadCLIP precisely detects abnormal region of different categories on two benchmarks, meanwhile, it also produces considerably low anomaly predictions on normal videos.

\section{Conclusion}
In this work, we propose a new paradigm named VadCLIP for weakly supervised video anomaly detection. To efficiently adapt the pre-trained knowledge and vision-language associations from frozen CLIP to WSVAD task, we first devise a LGT-Adapter to enhance the ability of temporal modeling, and then we design a series of prompt mechanisms to improve the adaptation of general knowledge to the specific task.
Finally we introduce the MIL-Align operation for facilitating the optimization of vision-language alignment under weak supervision. We empirically verify the effectiveness of VadCLIP through state-of-the-art performance and sufficient ablations on two WSVAD benchmarks. In future, we will continue to explore vision-language pre-trained knowledge and further devote to open-set VAD task. 

\section{Acknowledgments}
This work is supported by the National Natural Science Foundation of China (No. 62306240, U23B2013, U19B2037, 62301432, 62101453), China Postdoctoral Science Foundation (No. 2023TQ0272), National Key R\&D Program of China (No.2020AAA0106900), Shaanxi Provincial Key R\&D Program (No.2021KWZ-03), Natural Science Basic Research Program of Shaanxi (No. 2021JCW-03, 2023-JC-QN-0685), and the Fundamental Research Funds for the Central Universities (No. D5000220431).

\bibliography{aaai24}

\end{document}